\documentclass[10pt,twocolumn,a4paper]{esaAI}

\title{Optimizing Multi-Task Learning for Accurate Spacecraft Pose Estimation}

\def\authorEmail{francesco.evangelisti@aikospace.com}

\author[1]{Francesco Evangelisti \thanks{Corresponding author: \authorEmail}}
\author[1]{Francesco Rossi}
\author[1]{Tobia Giani}
\author[1]{Ilaria Bloise}
\author[1]{Mattia Varile}

\affil[1]{AIKO S.r.l., Turin, IT, \textit{www.aikospace.com}}

\begin{document}

% Creates the title and author list automatically for you!
\makeCustomtitle

% EDIT HERE
% Please add your abstract here, i.e., \begin{abstract}<Your abstract text>\end{abstract}
\begin{abstract}
Accurate satellite pose estimation is crucial for autonomous guidance, navigation, and control (GNC) systems in in-orbit servicing (IOS) missions. This paper explores the impact of different tasks within a multi-task learning (MTL) framework for satellite pose estimation using monocular images. By integrating tasks such as direct pose estimation, keypoint prediction, object localization, and segmentation into a single network, the study aims to evaluate the reciprocal influence between tasks by testing different multi-task configurations thanks to the modularity of the convolutional neural network (CNN) used in this work. The trends of mutual bias between the analyzed tasks are found by employing different weighting strategies to further test the robustness of the findings. A synthetic dataset was developed to train and test the MTL network.
Results indicate that direct pose estimation and heatmap-based pose estimation positively influence each other in general, while both the bounding box and segmentation tasks do not provide significant contributions and tend to degrade the overall estimation accuracy.
\end{abstract}

% EDIT HERE
% The main document. Please add your content as desired.
% We provide examples for adding figures, equations, and tables. Please stick to the style used in this template.

\section{Introduction}

Autonomous guidance, navigation, and control (GNC) systems are crucial for in-orbit servicing (IOS) missions, enabling tasks such as docking, repair, and refuelling. Accurate satellite pose estimation is essential for these operations. Traditional methods using multiple sensors like lidar and stereo cameras increase complexity and cost. This paper focuses on using monocular cameras for satellite pose estimation to streamline the process while maintaining high accuracy.

Recent advancements in artificial intelligence (AI), particularly convolutional neural networks (CNNs), have significantly improved computer vision tasks. However, monocular vision systems face limitations in scale ambiguity and depth perception, necessitating sophisticated algorithms for high precision \cite{bechini, linedet, pvspe}.

Multi-task learning (MTL) allows a single model to learn multiple related tasks simultaneously, leveraging shared representations to improve performance and efficiency \cite{mtl_human, yolo}. In satellite pose estimation, MTL can integrate direct pose estimation, keypoint detection, object detection, and segmentation into a unified framework, optimizing inference time and resource utilization.

A key development in this field is the Spacecraft Pose Network v2 (SPNv2)\cite{spnv2parkdamico}, which uses a multi-scale, multi-task CNN architecture to perform object detection, keypoint prediction, binary segmentation, and direct pose estimation. This approach has shown improved robustness and accuracy over previous methods. Moreover, SPNv2 has inspired the present research.

This paper explores the potential of MTL for satellite pose estimation using monocular cameras. By integrating various tasks into a single network, we aim to enhance the performance and efficiency of pose estimation systems for IOS missions. Our contributions include:
\begin{itemize}
\item Developing a synthetic dataset generation pipeline for training, validation, and testing.
\item Implementing a modular MTL network that outputs direct pose estimation, keypoint prediction, object detection, and segmentation from a single input image.
\item Evaluating the MTL network's performance to demonstrate its advantages over single-task learning and provide evidence for a task selection strategy to avoid negative and suboptimal setups.
\end{itemize}

\begin{figure*}[t]
    \centering
    \includegraphics[width=.85\textwidth]{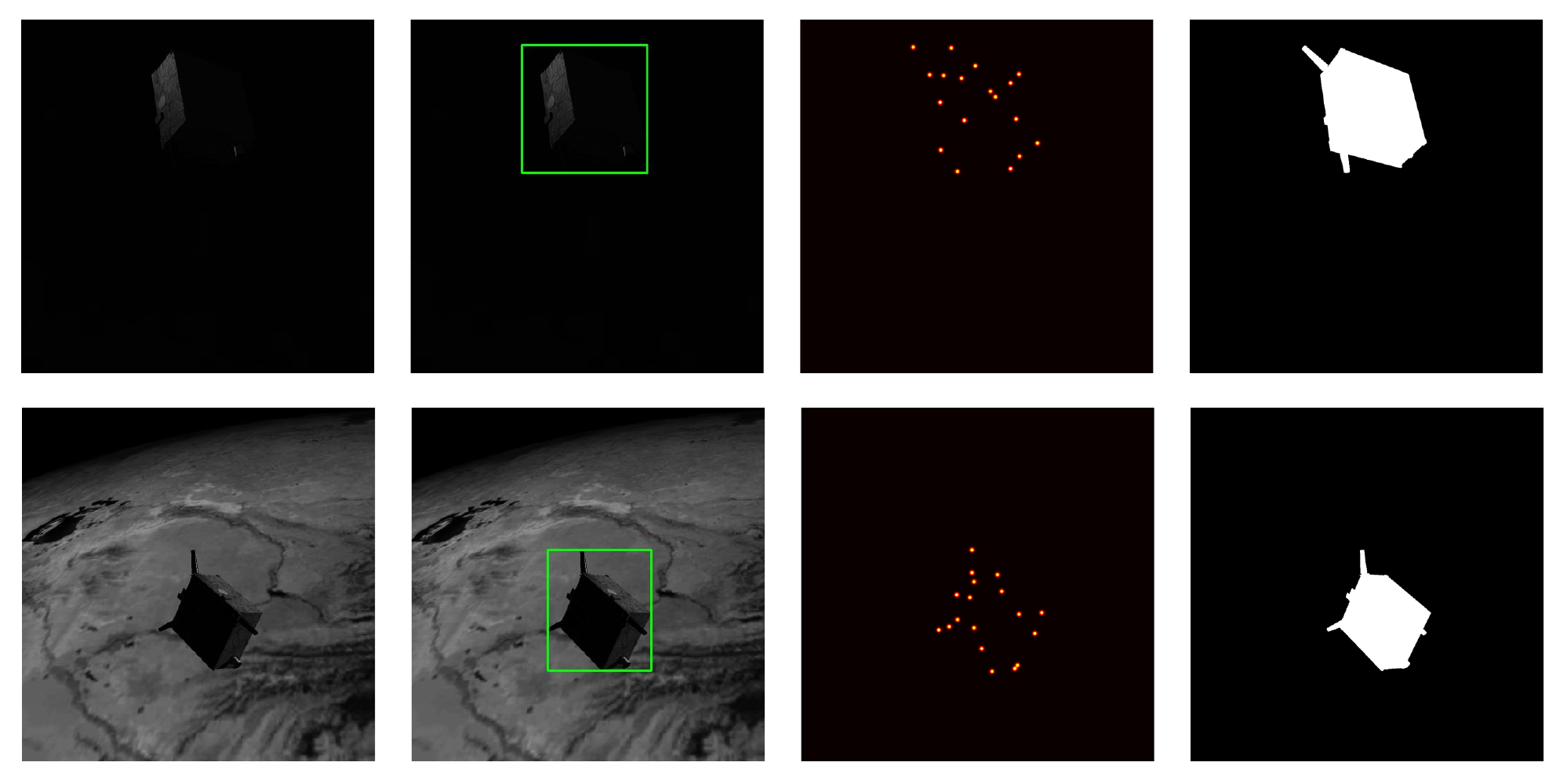}
	\caption{Two samples from our dataset. Each row corresponds to a sample. The first column contains the generated capture. The second column shows the bounding box. The third column displays the keypoints' heatmap. The fourth column contains the segmentation masks.}
	\label{fig:dataset_doublecolumn}
\end{figure*}

\section{Materials and Methods}
\subsection{Synthetic Data Generation}
Our dataset generation leverages a proprietary Unity-based setup, enabling the creation of both random and trajectory-based images of a satellite. This setup allows for a diverse range of scenarios and lighting conditions, providing a robust dataset essential for training and evaluating our multi-task learning (MTL) models.

For this work, we generated 40,000 images of the Tango satellite, split into 70\%-20\%-10\% for training, validation and testing. The dataset was randomized within a range of 1 to 25 meters to simulate various operational distances and orientations. The camera setup used for image generation included a resolution of 1024 × 1024 pixels, a focal length of 39.47 mm, and a pixel pitch of 5.86 µm/px. The horizontal and vertical fields of view were both set to 35.0 degrees. The images were rescaled to 512 × 512 pixels to make training and inference more feasible.

The dataset includes detailed metadata for each image, consisting of the pose of the satellite relative to the camera, the projected location of the 18 keypoints (differently from the 11 keypoints used in \cite{spnv2parkdamico}) distributed across the satellite on the camera frame, the bounding box of the satellite, and a binary segmentation mask. The keypoints in 2D locations are used to generate heatmaps used as ground-truths. The Tango satellite model is chosen as the target body in the foreground, while the background includes simulated space environments, providing a realistic and challenging dataset for our models. Some dataset samples from the test set are displayed in \cref{fig:dataset_doublecolumn}.

%Our pipeline is capable of automatically generating an unlimited number of images under specific conditions, including variations in lighting, camera setup, and noise addition. For this work, we employed a standard and balanced setup to ensure a comprehensive and representative dataset for our experiments. This approach replicates and extends the methodologies used in similar research, providing a solid foundation for training MTL models for satellite pose estimation.

\subsection{Model Architecture and Setup}
The developed network architecture is designed to handle multiple tasks from a single input image: direct pose estimation, indirect pose estimation via keypoint prediction, bounding box estimation, and segmentation mask.

The network utilizes an EfficientNet \cite{efficientnet} backbone for feature extraction and a Bidirectional Feature Pyramid Network (BiFPN) to enhance feature representation across scales. The heads attached to the backbone are EfficientPose \cite{efficientpose} for object localization and direct pose estimation, a heatmap head for keypoints prediction, and a segmentation map head for target segmentation.

A key feature of our network implementation is its modularity, which makes it fully configurable. This design allows us to easily activate or deactivate specific tasks, enabling flexible experimentation and optimization. This modular approach is critical to fine-tuning the network and achieving optimal performance for specific tasks. The development framework utilized was PyTorch 2.0 \cite{ansel2024pytorch}.

\begin{figure}[h]
    \centering
    \includegraphics[width=.9\columnwidth]{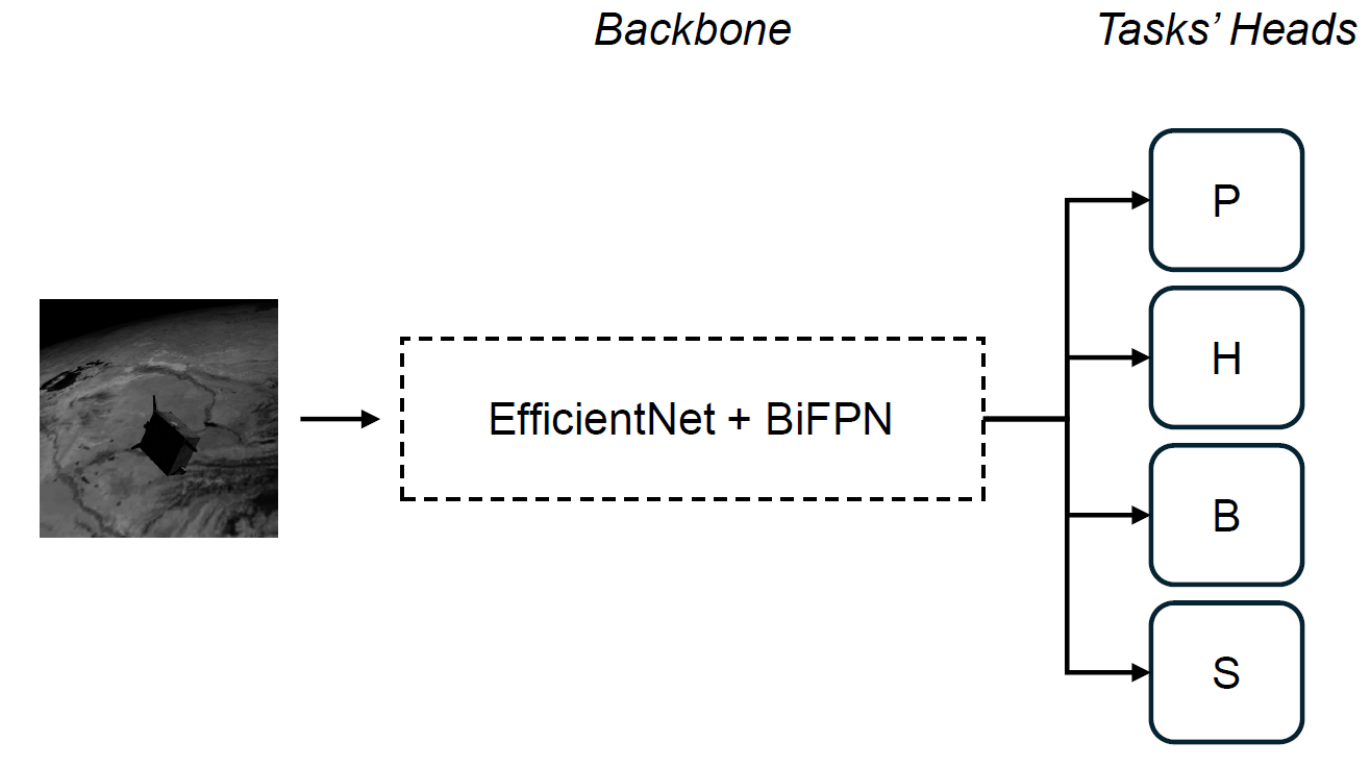}
	\caption{Simplified visualization of the proposed CNN MTL architecture. P, H, B, and S respectively represent the heads for direct pose estimation, keypoints' heatmaps prediction, bounding box prediction, and segmentation tasks.
	}
	\label{fig:CNN_architecture}
\end{figure}

The network tasks are labelled as follows: direct pose estimation (\textbf{P}), indirect pose estimation or heatmap-based pose estimation (\textbf{H}), bounding box estimation (\textbf{B}), and segmentation estimation (\textbf{S}). The indirect pose estimation is obtained by solving the PnP problem by exploiting the predicted keypoints' heatmaps \cite{pnp}. A representation of the network is presented in \cref{fig:CNN_architecture}.

Managing multiple outputs in a MTL network requires effective weighting strategies to balance the importance of each task. This is particularly relevant when tasks are heterogeneous and qualitatively different.

Our setup allows for both manual and automatic task weighting strategies. By default, tasks are assigned equal weighting (\textbf{EW}), but other strategies can be employed, including:
\begin{itemize}
\item Random Loss Weighting (\textbf{RLW}), which assigns random weights to tasks during each training iteration \cite{ws_rlw};
\item Dynamic Weight Average (\textbf{DWA}), which adjusts task weights dynamically based on the training loss trends \cite{ws_dwa};
\item Gradient Normalization (\textbf{GradNorm}), which normalizes gradients to ensure balanced learning across tasks \cite{ws_gradnorm}.
\end{itemize}

In this study, we investigate whether auxiliary tasks aid the primary task of pose estimation, crucial for satellite navigation.

We first train the network with only direct pose estimation (P) as the primary task. Next, we train with all tasks included (PHBS). To assess the impact of each auxiliary task, we use a leave-one-out strategy: excluding heatmap-based pose estimation (PBS), bounding box estimation (PHS), and segmentation estimation (PHB).

We also test configurations by adding each auxiliary task individually to the primary task (P): heatmap-based pose estimation (PH), bounding box estimation (PB), and segmentation estimation (PS).

Finally, to understand the role of auxiliary tasks without direct pose estimation (P), we use heatmap-based pose estimation (H) as the primary task and test combinations with bounding box (HB), segmentation (HS), and both tasks (HSB).

This approach allows us to evaluate the effectiveness of auxiliary tasks in enhancing the primary task of pose estimation.

For evaluating the performance of our pose estimation model, we use the following metrics:
\begin{itemize}
    \item \textbf{Translation Error}: Measures the Euclidean distance between the estimated translation vector \(\hat{t}\) and the ground truth translation vector \(t\):
    \[
    E_T = \| \hat{t} - t \|
    \]
    \item \textbf{Rotation Error}: Quantifies the difference between the estimated rotation matrix \(\hat{R}\) and the ground truth rotation matrix \(R\):
    \[
    E_R = \arccos \left( \frac{\text{trace}(\hat{R}R^T) - 1}{2} \right)
    \]
    \item \textbf{SPEED Score}: A composite metric evaluating overall pose estimation performance by considering both rotation error and normalized translation error:
    \[
    \text{\textbf{SPEED score}} = E_{\text{pose}} = E_R(\hat{R}, R) + \frac{E_T(\hat{t}, t)}{\|t\|}
    \]
\end{itemize}

% LOSSES description
The SPEED score is also used as the loss for the direct pose estimation task. The object localization task is trained by the means of the \textit{Complete Intersection over Union} (C-IoU) loss, while both the keypoints' heatmaps and target segmentation tasks are associated with a \textit{pixel-wise Mean Squared Error} (MSE) loss.

All the experiments were conducted with a consistent number of training epochs, identical hyperparameters (as detailed in \cref{tab:hyperparameters}), with the same batch size (BS) and the same learning rate (LR) decay schedule.

\begin{table}[h]\renewcommand{\arraystretch}{1.2}
\begin{center}
\begin{tabular}{ c | c | c | c | c } 
\hline\hline
Epochs & BS & LR & LR steps & LR factor \\
\hline\hline
40 & 16 & $5\times 10^{-4}$ & 75\% - 90 \% & $1\times 10^{-1}$ \\
\hline\hline
\end{tabular}
\caption{Training hyperparameters.}
\label{tab:hyperparameters}
\end{center}
\end{table}

The network's backbone was scaled down to create a lightweight version, allowing for a high number of experiments. This resulted in the selection of the smallest version of EfficientNet, \textit{EfficientNet-B0}. The number of parameters characterizing the backbone and the prediction heads involved in the experiments are summarized in \cref{tab:network_params}.

\begin{table}[h]\renewcommand{\arraystretch}{1.2}
\begin{center}
\begin{tabular}{ c | c } 
\hline\hline
\textbf{Network block} & \textbf{Number of parameters} \\
\hline\hline
EfficientNet-B0\textsuperscript{*} & 3,824,772 \\
P & 90,116 \\
H & 48,082 \\
B & 18,852 \\
D & 45,889 \\
\hline
Tot. & 4,024,711 \\
\hline\hline
\end{tabular}
\caption{Number of network parameters. The block identified by ({*}) is the network's backbone; the other blocks are prediction heads.}
\label{tab:network_params}
\end{center}
\end{table}

Since the focus is exclusively on pose estimation, the model size could be reduced for deployment by eliminating all auxiliary heads after they are used for training. Furthermore, these findings could be investigated using alternative, more compact backbones or architectures, potentially reducing the parameter count while preserving performance.

The training sessions were conducted on an NVIDIA RTX A6000 GPU. The chosen number of epochs represents a balance between training duration and the achieved performance in pose estimation. Additionally, the optimization of the network's total loss begins to plateau after 40 epochs, thus further improvements are not considered beneficial to the objectives of this study.

\section{Results}
In this section, we present the results of our experiments trained and tested to evaluate the effectiveness of our MTL network for spacecraft pose estimation. Our primary objective was to determine whether auxiliary tasks can improve the accuracy of the pose estimation tasks.
We begin by evaluating the direct (P) and indirect (H) pose estimation tasks in a single-task setup. This configuration serves as our baseline for comparing the impact of including auxiliary tasks. The results calculated on the test-set are summarized in the form of SPEED score in \cref{tab:results}.

\begin{figure}[h]
    \centering
    \includegraphics[width=.9\columnwidth]{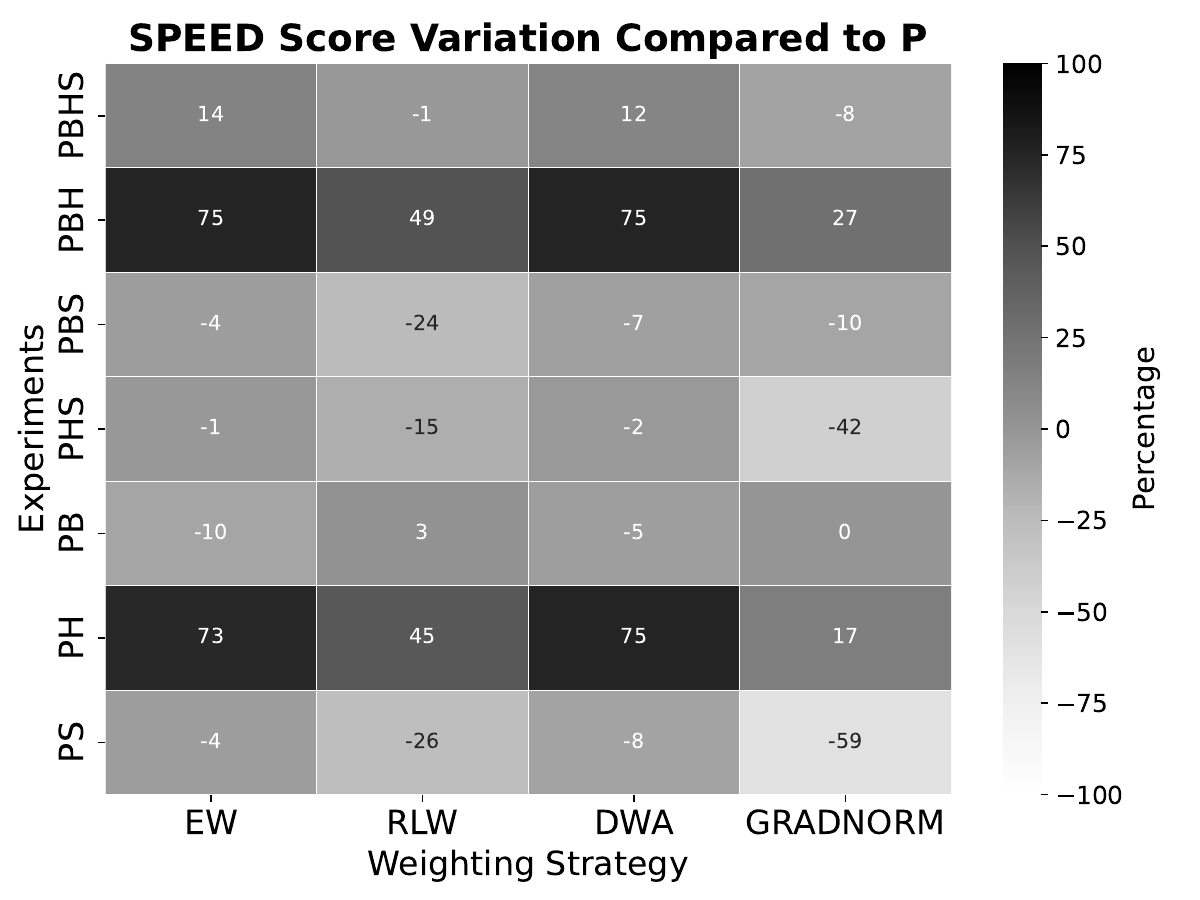}
    \caption{Percentual change in SPEED score from direct pose estimation compared to the single task network P performance. A positive change means a reduction in the score (the lower the better).
	}
	\label{fig:Benchmark_P}
\end{figure}

\begin{table}[h]\renewcommand{\arraystretch}{1.2}
\begin{center}
\begin{tabular}{ c || c | c } 
\hline\hline
Model & {Median} & {IQR}\\
\hline\hline
P & $0.052$ & $0.046$ \\
H & $0.042$ & $0.035$ \\
\hline\hline
\end{tabular}
\caption{SPEED scores for single task networks P and H.}
\label{tab:results}
\end{center}
\end{table}
We tested the direct pose estimation task (P) with various combinations of auxiliary tasks (H, B, S) and different weighting strategies. The configurations include complete multi-task training (PHBS), leave-one-out strategies (PBS, PHS, PHB), and single auxiliary task addition (PH, PB, PS). The results are presented as the percentage improvement in SPEED score relative to the single-task baseline in \cref{fig:Benchmark_P}. Percentage changes from baseline results in pose estimation performance provide an immediate and straightforward way to compare the efficacy of different configurations and are more convenient to read than absolute numbers for the SPEED score to highlight variations.

\begin{figure}[h]
    \centering
    \includegraphics[width=.9\columnwidth]{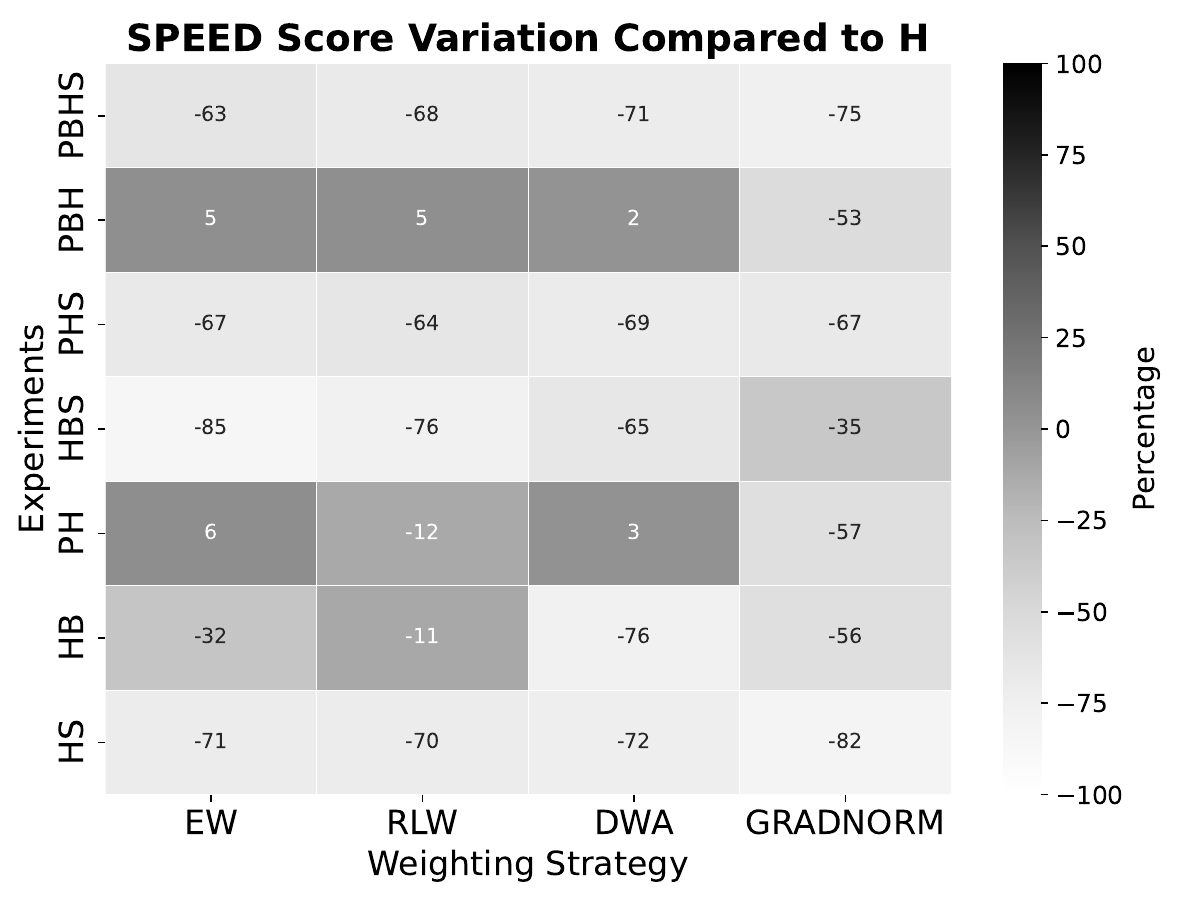}
     \caption{Percentual change in SPEED score from indirect pose estimation compared to the single task network H performance. A positive change means a reduction in the score (the lower the better).
	}
	\label{fig:Benchmark_H}
\end{figure}

In \cref{fig:Benchmark_H}, the same is shown for the indirect pose estimation (H).

\begin{figure}[h]
    \centering
    \includegraphics[width=0.9\columnwidth]{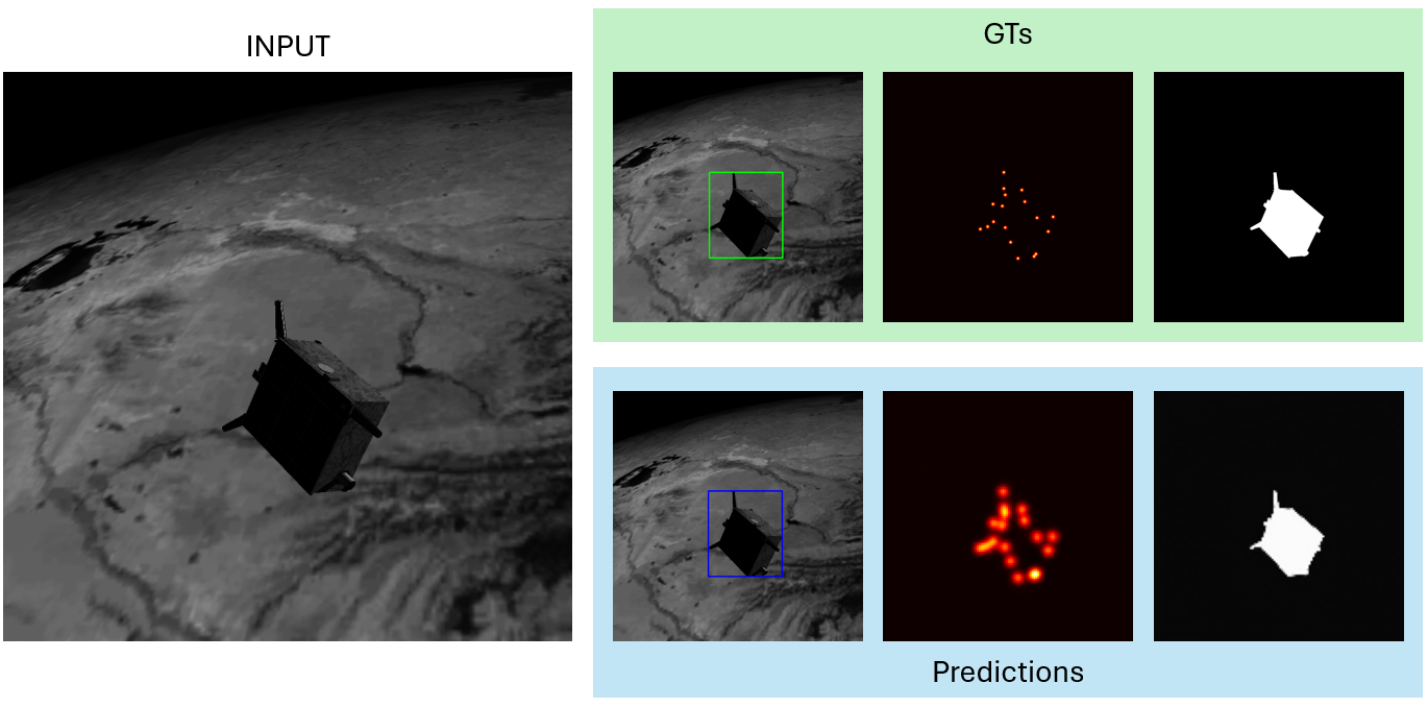}
	\caption{Inference results on a test sample. The predicted bounding box, heatmaps and segmentation are highlighted in the bottom row, while the relative ground truths are in the upper row.}
	\label{fig:PBHS_predictions}
\end{figure}

For the sake of completeness, the qualitative results from the inference on a test image of the PBHS model trained through the equal weighting strategy are displayed in \cref{fig:PBHS_predictions}.

\section{Conclusion and Discussion}

The comparative results between single-task direct pose estimation (P) and multi-task learning (MTL) solutions indicate that the indirect pose estimation (H) task is beneficial to the direct one (P). In contrast, bounding box estimation (B) does not significantly impact the results, and segmentation estimation (S) negatively affects the overall pose estimation accuracy.  This negative influence is likely due to the differing nature of the segmentation task compared to the others, which may translate to a different scale for the selected loss.

These findings hold across various weighting strategies, suggesting that different strategies do not substantially alter the overall trend. Among the evaluated methods, Equal Weighting (EW) and Dynamic Weight Average (DWA) were identified as the most effective.

In the context of heatmap-based (indirect) pose estimation (H), the presence of the segmentation task (S) also resulted in degraded performance. Furthermore, H did not generally receive positive contributions from other tasks, except for direct pose estimation (P). This trend was consistent across different weighting strategies, reinforcing the robustness of the results.

Overall, GradNorm was found to be the least effective weighting strategy in our evaluations.

While the effects of the experiments on the indirect pose estimation are limited, revealing some kind of knowledge saturation for the keypoints' heatmap regression task (H), the smallest model to achieve the best SPEED score from the direct estimation task (P) is the PH configuration, trained with the DWA strategy. For this model, a 75\% performance boost with respect to the baseline model (P only) leads to a median SPEED score of 0.013. This score value is low when compared to results presented in \cite{spnv2parkdamico} using the same backbone. This may be partly attributed to the data on which the model is trained, but it is crucial to acknowledge how exploring different network configurations together with diverse weighting strategies led to achieving a 75\% enhancement on the direct pose estimation task.

Future work will test the repeatability of the experiments using more domain-representative simulated data for training and testing.

It is to be noticed that the results presented are contingent upon both the global and task-specific configurations of the network; variations in individual task losses within the MTL framework can yield differing outcomes. Consequently, these results and experimental setups should be considered as a baseline for upcoming advancements.

Future research will also focus on exploring alternative weighting strategies to optimize the synergy and mutual information embedded within the metadata.
Additionally, it can be important to assess the impact of additional types of metadata (e.g., other ground-truths) to enhance the performance and robustness of the multi-task learning framework.

\printbibliography

@article{spnv2parkdamico,
title={Robust multi-task learning and online refinement for spacecraft pose estimation across domain gap},
volume={73},
ISSN={0273-1177},
url={http://dx.doi.org/10.1016/j.asr.2023.03.036},
DOI={10.1016/j.asr.2023.03.036},
number={11},
journal={Advances in Space Research},
publisher={Elsevier BV},
author={Park, Tae Ha and D’Amico, Simone},
year={2024},
month=jun, pages={5726–5740} }

@inproceedings{linedet,
title={Towards light-weight and real-time line segment detection},
author={Gu, Geonmo and Ko, Byungsoo and Go, SeoungHyun and Lee, Sung-Hyun and Lee, Jingeun and Shin, Minchul},
booktitle={Proceedings of the AAAI Conference on Artificial Intelligence},
volume={36},
number={1},
pages={726--734},
year={2022}
}

@article{bechini,
title = {Robust spacecraft relative pose estimation via CNN-aided line segments detection in monocular images},
journal = {Acta Astronautica},
volume = {215},
pages = {20-43},
year = {2024},
issn = {0094-5765},
doi = {https://doi.org/10.1016/j.actaastro.2023.11.049},
url = {https://www.sciencedirect.com/science/article/pii/S0094576523006185},
author = {Michele Bechini and Geonmo Gu and Paolo Lunghi and Michèle Lavagna}
}

@article{yolo,
title={You only look at once for real-time and generic multi-task},
author={Wang, Jiayuan and Wu, QM Jonathan and Zhang, Ning},
journal={IEEE Transactions on Vehicular Technology},
year={2024},
publisher={IEEE}
}

@article{mtl_human,
  title={A Real-Time Multi-Task Learning System for Joint Detection of Face, Facial Landmark and Head Pose},
  author={Wu, Qingtian and Zhang, Liming},
  journal={arXiv preprint arXiv:2309.11773},
  year={2023}
}

@article{pvspe,
  title={PVSPE: A Pyramid Vision Multitask Transformer Network for Spacecraft Pose Estimation},
  author={Yang, Hong and Xiao, Xueming and Yao, Meibao and Xiong, Yonggang and Cui, Hutao and Fu, Yuegang},
  journal={Advances in Space Research},
  year={2024},
  publisher={Elsevier}
}

@article{ws_rlw,
  title={Reasonable effectiveness of random weighting: A litmus test for multi-task learning},
  author={Lin, Baijiong and Ye, Feiyang and Zhang, Yu and Tsang, Ivor W},
  journal={arXiv preprint arXiv:2111.10603},
  year={2021}
}

@inproceedings{ws_dwa,
  title={End-to-end multi-task learning with attention},
  author={Liu, Shikun and Johns, Edward and Davison, Andrew J},
  booktitle={Proceedings of the IEEE/CVF conference on computer vision and pattern recognition},
  pages={1871--1880},
  year={2019}
}

@inproceedings{ws_gradnorm,
  title={Gradnorm: Gradient normalization for adaptive loss balancing in deep multitask networks},
  author={Chen, Zhao and Badrinarayanan, Vijay and Lee, Chen-Yu and Rabinovich, Andrew},
  booktitle={International conference on machine learning},
  pages={794--803},
  year={2018},
  organization={PMLR}
}

@inproceedings{pnp,
  title={A direct least-squares (DLS) method for PnP},
  author={Hesch, Joel A and Roumeliotis, Stergios I},
  booktitle={2011 International Conference on Computer Vision},
  pages={383--390},
  year={2011},
  organization={IEEE}
}

@inproceedings{efficientnet,
  title={Efficientnet: Rethinking model scaling for convolutional neural networks},
  author={Tan, Mingxing and Le, Quoc},
  booktitle={International conference on machine learning},
  pages={6105--6114},
  year={2019},
  organization={PMLR}
}

@article{efficientpose,
  title={EfficientPose: An efficient, accurate and scalable end-to-end 6D multi object pose estimation approach},
  author={Bukschat, Yannick and Vetter, Marcus},
  journal={arXiv preprint arXiv:2011.04307},
  year={2020}
}

@inproceedings{ansel2024pytorch,
  title={PyTorch 2: Faster Machine Learning Through Dynamic Python Bytecode Transformation and Graph Compilation},
  author={Ansel, Jason and Yang, Edward and He, Horace and Gimelshein, Natalia and Jain, Animesh and Voznesensky, Michael and Bao, Bin and Bell, Peter and Berard, David and Burovski, Evgeni and others},
  booktitle={Proceedings of the 29th ACM International Conference on Architectural Support for Programming Languages and Operating Systems, Volume 2},
  pages={929--947},
  year={2024}
}
\addcontentsline{toc}{section}{References}

\end{document}